\documentclass[10pt,twocolumn,letterpaper]{article}
\pdfoutput=1
\usepackage{3dv}
\usepackage{times}
\usepackage{epsfig}
\usepackage{graphicx}
\usepackage{caption}
\usepackage{subcaption}
\usepackage{float}

\usepackage[utf8]{inputenc}
\usepackage[T1]{fontenc} 
\usepackage{caption} 
\usepackage[subrefformat=parens,labelformat=parens]{subcaption}  
\usepackage{multirow} 

\makeatletter
\@namedef{ver@everyshi.sty}{}
\makeatother
\usepackage{pgfplots}
\usepackage{pgf,tikz}

\usepackage{listings}
\usepackage{color} %red, green, blue, yellow, cyan, magenta, black, white
\definecolor{mygreen}{RGB}{28,172,0} % color values Red, Green, Blue
\definecolor{mylilas}{RGB}{170,55,241}

\usepackage{amsmath,amssymb,amsthm}

\newcommand{\norm}[1]{\left\lVert#1\right\rVert}

\newcommand{\Rr}{\mathbb{R}}
\newcommand{\Xx}{\mathcal{X}}

\newcommand{\bb}[1]{\mathbf{#1}}

\usepackage[pagebackref=true,breaklinks=true,letterpaper=true,colorlinks,bookmarks=false]{hyperref}

\threedvfinalcopy % *** Uncomment this line for the final submission

\def\threedvPaperID{231} % *** Enter the 3DV Paper ID here

\begin{document}

\lstset{language=Matlab,%
    %basicstyle=\color{red},
    breaklines=true,%
    morekeywords={matlab2tikz},
    keywordstyle=\color{blue},%
    morekeywords=[2]{1}, keywordstyle=[2]{\color{black}},
    identifierstyle=\color{black},%
    stringstyle=\color{mylilas},
    commentstyle=\color{mygreen},%
    showstringspaces=false,%without this there will be a symbol in the places where there is a space
    numbers=left,%
    numberstyle={\tiny \color{black}},% size of the numbers
    numbersep=9pt, % this defines how far the numbers are from the text
    emph=[1]{for,end,break},emphstyle=[1]\color{red}, %some words to emphasise
    %emph=[2]{word1,word2}, emphstyle=[2]{style},    
}

\pagenumbering{gobble}

%%%%%%%%% TITLE
\title{Unsupervised Dense Shape Correspondence using Heat Kernels}

\author{Mehmet Ayg\"{u}n \\
\and
Zorah L{\"a}hner \\
{Technical University of Munich}\\{\tt\small \{mehmet.ayguen,zorah.laehner,cremers\}@tum.de}
\and Daniel Cremers 
}

\maketitle

\begin{figure*}[h]
     \centering
     \begin{subfigure}[b]{0.16\textwidth}
         \centering
         \includegraphics[height=0.17\textheight]{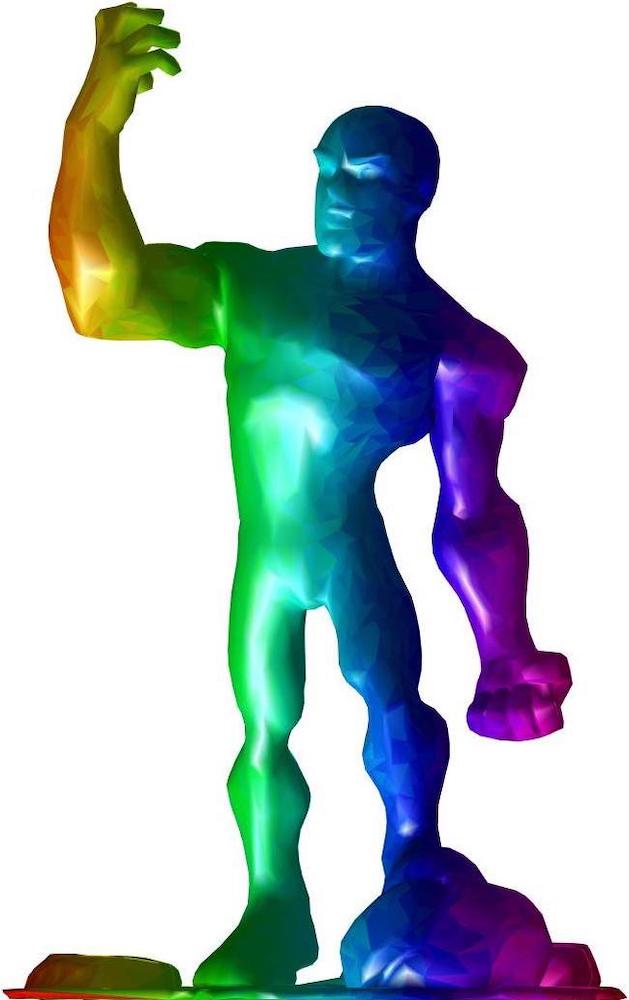}
         \caption{Reference}
         \label{fig:single_ref}
     \end{subfigure}
     \hfill
          \begin{subfigure}[b]{0.16\textwidth}
         \centering
         \includegraphics[width=\textwidth]{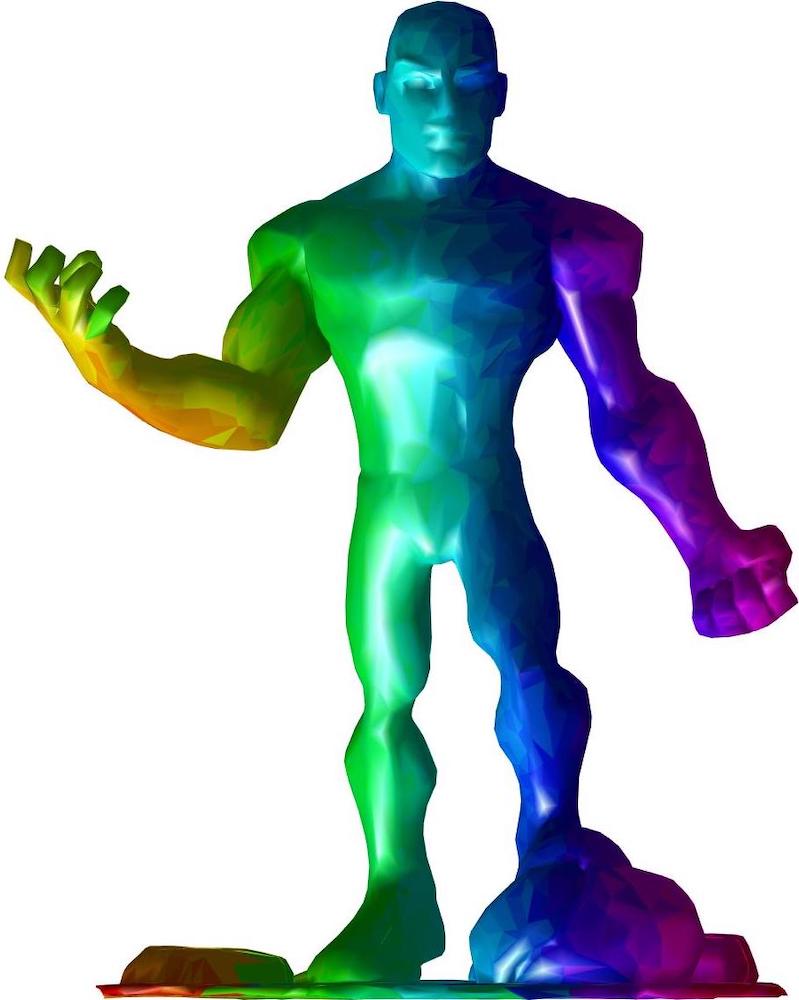}
         \caption{Heat Kernel}
         \label{fig:single_heat}
     \end{subfigure}
     \hfill
          \begin{subfigure}[b]{0.16\textwidth}
         \centering
         \includegraphics[width=\textwidth]{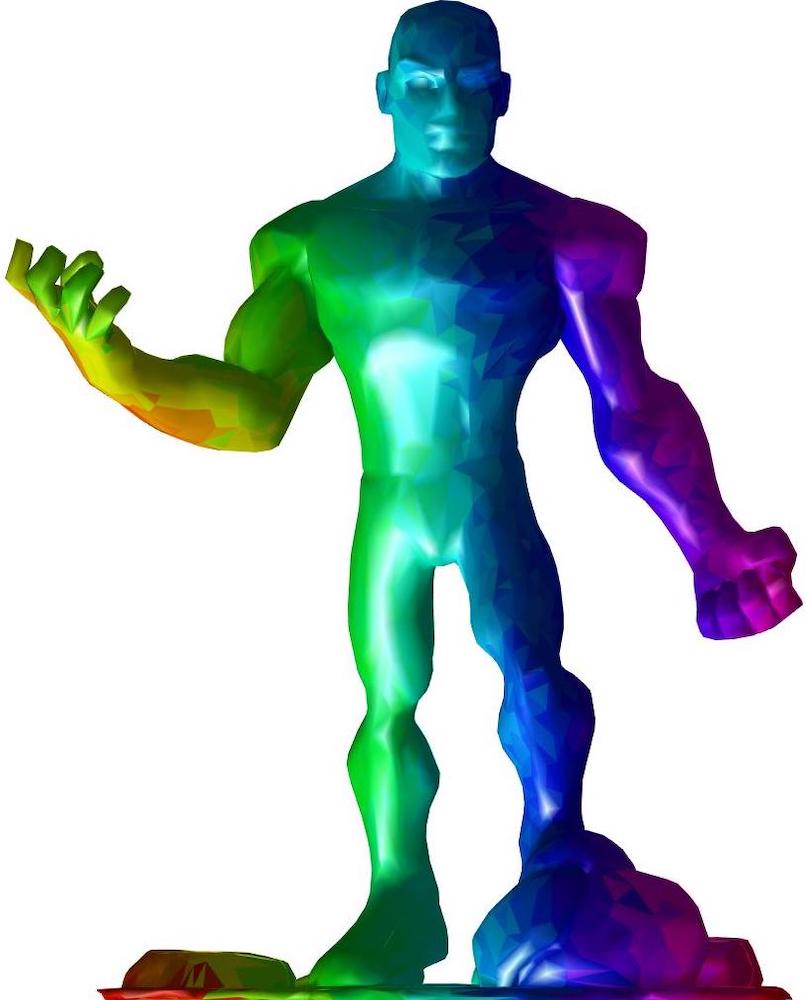}
         \caption{Geodesic Dist\cite{halimi2019unsupervised}}
         \label{fig:single_uns}
     \end{subfigure}
     \hfill
          \begin{subfigure}[b]{0.16\textwidth}
         \centering
         \includegraphics[width=\textwidth]{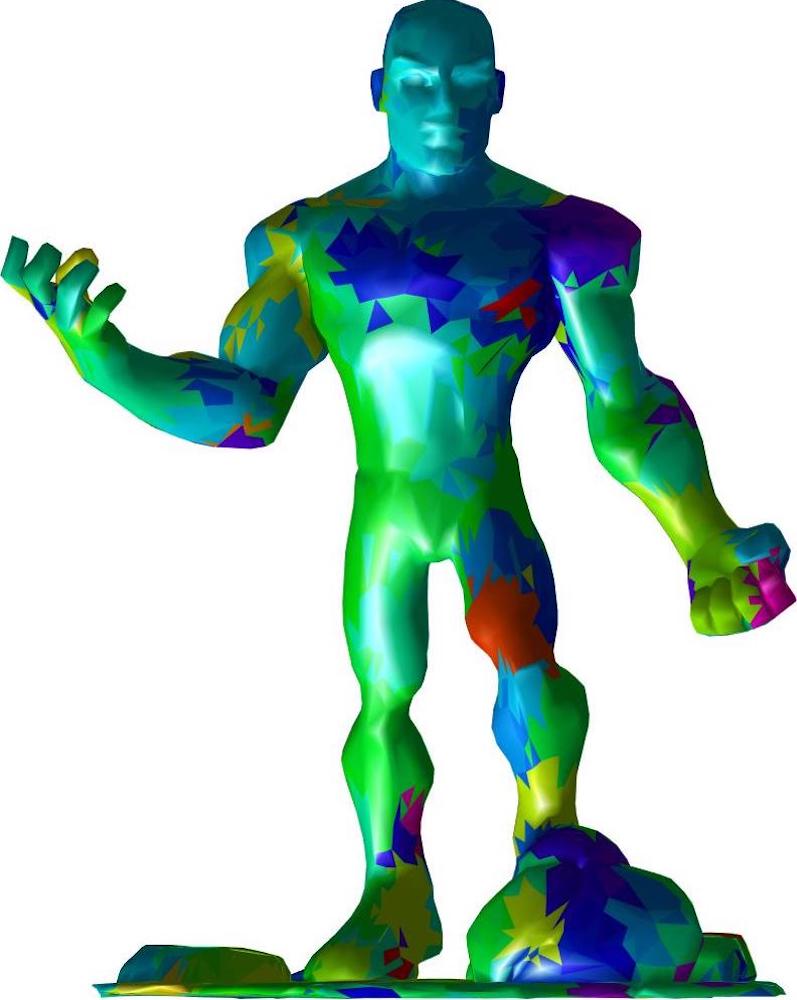}
         \caption{Supervised \cite{litany2017deep}}
         \label{fig:single_sup}
     \end{subfigure}
     \hfill
               \begin{subfigure}[b]{0.16\textwidth}
         \centering
         \includegraphics[width=\textwidth]{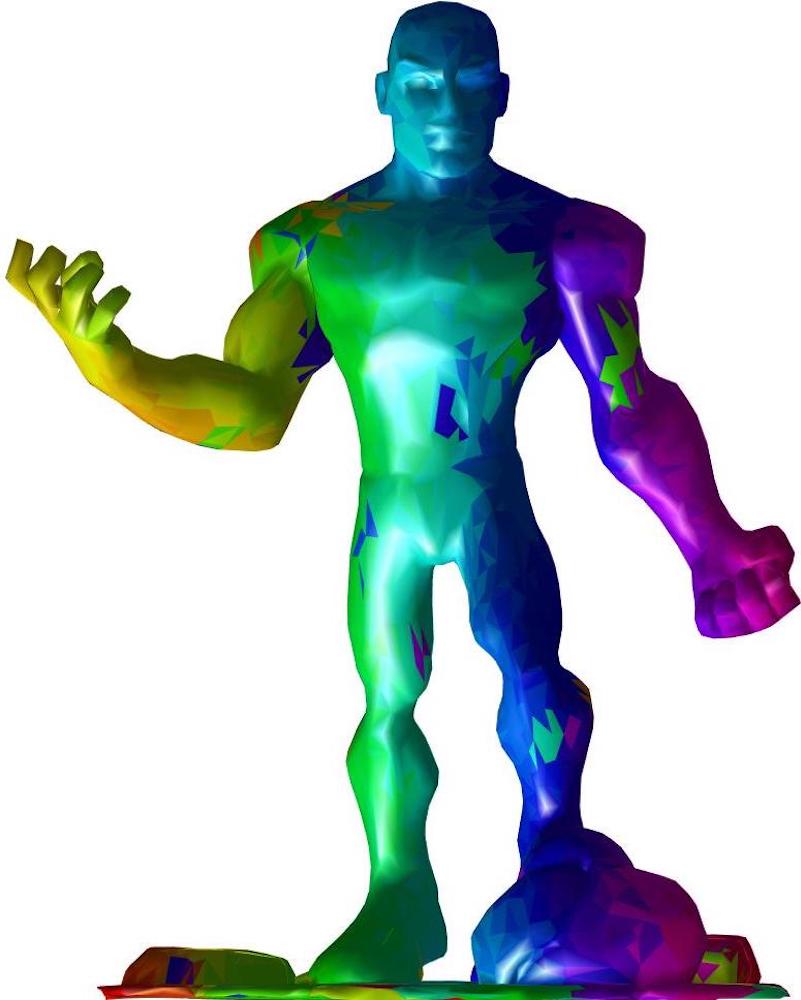}
         \caption{SGMDS \cite{aflalo2016spectral}}
         \label{fig:single_sgmds}
     \end{subfigure}
     \hfill
 \begin{subfigure}[b]{0.16\textwidth}
         \centering
         \includegraphics[width=\textwidth]{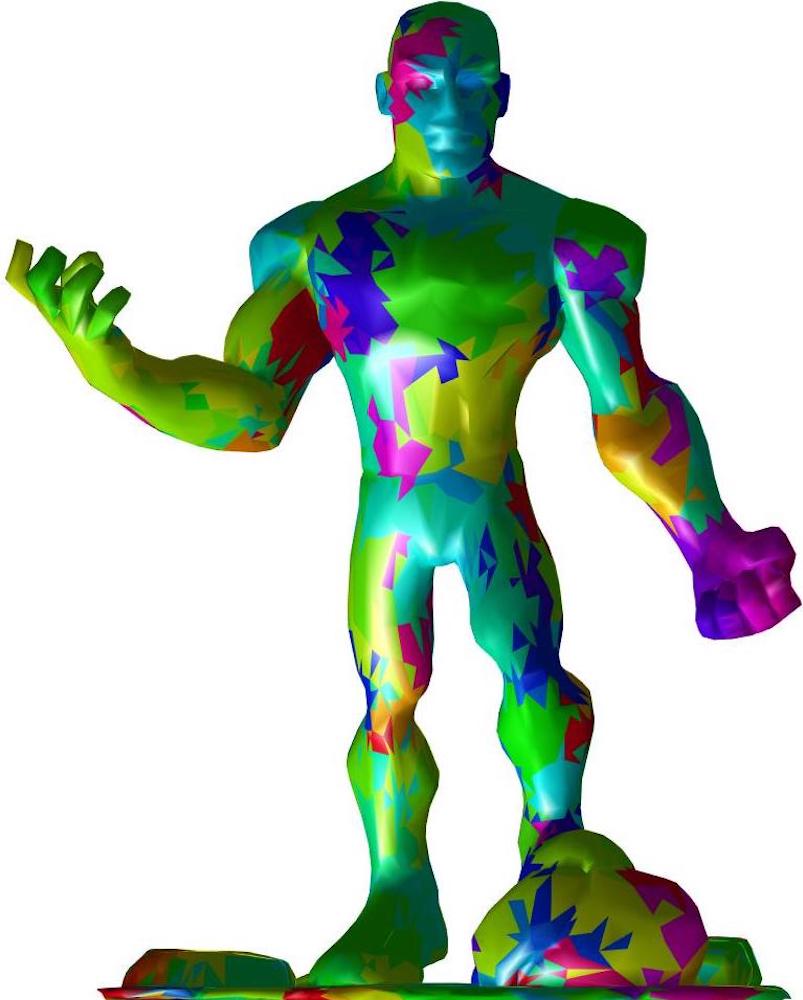}
         \caption{F. Map\cite{ovsjanikov2012functional}}
         \label{fig:single_fmap}
     \end{subfigure}
        \caption{Dense correspondence between two shapes optimized only on a single pair of shapes. Since there is no ground-truth available, our method (\ref{fig:single_heat}) and \ref{fig:single_uns} are trained in an unsupervised manner. Moreover, while \cite{halimi2019unsupervised} took $15$ minutes to optimize, our method took just $1$ min. We did not employ post-processing method to any of the results.}
        \label{fig:single_exp}
\end{figure*}
\pagestyle{plain}

%%%%%%%%% ABSTRACT
\begin{abstract}
In this work, we propose an unsupervised method for learning dense correspondences between shapes using a recent deep functional map framework. Instead of depending on ground-truth correspondences or the computationally expensive geodesic distances, we use heat kernels. These can be computed quickly during training as the supervisor signal. Moreover, we propose a curriculum learning strategy using different heat diffusion times which provide different levels of difficulty during optimization without any sampling mechanism or hard example mining. We present the results of our method on different benchmarks which have various challenges like partiality, topological noise and different connectivity.
\end{abstract}

%%%%%%%%% BODY TEXT
\section{Introduction}
\label{sec:intro}

With the growing market of 3D scanners and capturing systems, applications around 3D scanning are becoming more popular. 
One of the main building blocks of these applications in the domain of Augmented Reality/Virtual Reality is aligning scans and calculating correspondences between them.
Correspondences are necessary to apply detected user motion to an avatar or transfer properties of objects in the virtual world. 
Finding correspondences between shapes is a difficult problem in computer vision and graphics. Especially challenging cases are non-rigid motion, scanning noise (e.g. partiality, topological noise), different resolutions or connectivity between source and target shapes. 

In recent years, various approaches were suggested for the non-rigid shape correspondence problem. 
Similar to the development in image processing, the first approaches chose to solve a descriptor matching problem \cite{aubry2011wave, bronstein2010scale, rustamov2007laplace, sun2009concise, tombari2010unique}. 
However, designing descriptors which are invariant to various kinds of deformations and noise, as well as sufficiently discriminative is not easy. 
Nowadays better results can be achieved with learned descriptors instead of hand-crafted ones \cite{boscaini2015learning, boscaini2016anisotropic, litman2013learning, cosmo2016matching, corman2014supervised}. 

Recent methods have adapted to include other parts of correspondence pipelines into the learning process instead of just descriptors.
With the functional map framework \cite{ovsjanikov2012functional}, shape matching can be formulated as a learning problem in function space with the aim of minimizing distance distortion \cite{litany2017deep}. 
This approach uses ground truth correspondences for training the network in order to determine the distortion. 
Later, \cite{halimi2019unsupervised, roufosse2019unsupervised} introduced networks which can be trained in an unsupervised way. 
This is a big advantage, because ground-truth correspondences only exist for a very limited amount of shape collections.
While Halimi \etal \cite{halimi2019unsupervised} used the geodesic distance matrix as the supervisor signal, Roufosse \etal \cite{roufosse2019unsupervised} used constraints on functional maps to optimize their deep networks, but both are based on the network architecture introduced in \cite{litany2017deep}. 
Not relying on ground-truth correspondences eliminates the need for labelled data, and Halimi \etal \cite{halimi2019unsupervised} used geodesic distance matrices for this purpose.
However, geodesic distances are computationally expensive, not stable on degenerated meshes, and sensitive to topological noise on the surface. 
As a solution, we suggest to use heat kernels instead of geodesic distance matrices for unsupervised learning. 

In this paper, we argue that using heat kernels is almost always beneficial to geodesic distances.
Heat kernel are fast to compute, more stable on noisy meshes, have advantageous properties during optimization \cite{vestner2017efficient}, and their time parameter offers control on the amount of localization.
A heat kernel with a large time parameter includes information on the entire shape whereas a small time parameter is limited to only the close vicinity around one point.
We use this property to design a curriculum learning strategy for the training process and show that it benefits finding a good optimum. 
This can be interpreted as a coarse-to-fine optimization strategy similar to approaches introduced in \cite{vestner2017efficient, melzi2019zoomout}. However, in our proposed curriculum learning the coarse-to-fine optimization is only needed during learning and not at inference time. This makes the final operations even more efficient. 

Our main contribution can be summarized as \textbf{i)} showing that heat kernels, which are extremely cheap to compute compared to geodesic distances, can be used as a supervision signal for learning shape correspondence using the deep functional framework and produce even better results,
\textbf{ii)} using different time parameters for heat diffusion allows us to design a curriculum learning approach which boost the correspondence accuracy on different benchmarks.

The rest of the paper is organized as follows: In the Sec.~\ref{sec:rl} we discuss previous approaches and the latest deep learning based shape correspondence methods. 
In Sec.~\ref{sec:background} we give theoretical background about the shape representation that we used, heat kernels and (deep) functional maps, which are fundamental blocks of our method. 
Sec.~\ref{sec:method} explains how we used heat kernels as a supervisor signal for training and introduces our curriculum learning approach. In Sec.~\ref{sec:exp} we present our results on a variety of benchmarks. 

\section{Related Work}
\label{sec:rl}
Finding correspondences between two shapes is often formulated as a minimization problem using similarity measures between points. This can be done via matching of point-wise descriptors. These can be hand-crafted \cite{aubry2011wave, bronstein2010scale, rustamov2007laplace, sun2009concise, tombari2010unique} or learned \cite{boscaini2015learning, boscaini2016anisotropic, litman2013learning, cosmo2016matching, corman2014supervised}. 
A similar direction is to match pair-wise descriptors like distances to find the optimal matching \cite{coifman2005geometric, memoli2005theoretical, vestner2017efficient}.
%The In general, the first step is extracting features for all vertices, then trying to matching with other shape, while minimizing distance like geodesic distance.  

Functional maps \cite{ovsjanikov2012functional} had a big impact by modelling the correspondence problem as a mapping between functions on the surface instead of the points of the surface directly.
The main advantages is that using the Laplace-Beltrami eigenbasis for representing these functions reduces the dimensionality of the problem drastically. 
The idea of functional maps was refined for various applications \cite{pokrass2013sparse, rodola2017partial, huang2014functional, kovnatsky2015functional,aflalo2016spectral, eynard2016coupled, litany2016non, litany2017fully, nogneng2017informative} and recently adapted in deep learning frameworks \cite{litany2017deep, halimi2019unsupervised, roufosse2019unsupervised}. 
Different from descriptor based learning approaches, the optimization focuses on functions represented in a limited basis which is low dimensional compared to the number of vertices. 
After obtaining functional map between shapes, correspondences between vertices can be extracted in various ways \cite{ovsjanikov2012functional, rodola15cpd}.  

\subsection{Deep Learning Based Shape Correspondence}

Monti \etal\cite{monti2017geometric} proposed an unified framework for extending convolutional neural networks on non-euclidean data such as manifolds and graphs. The proposed network learns task specific local features. They modelled dense correspondence as a classification task.

Deep Functional Maps \cite{litany2017deep} introduced a framework for learning dense shape correspondence between 3D shapes using deep learning and functional maps. 
Instead of depending on hand-crafted local descriptors such as SHOT \cite{salti2014shot}, the optimal descriptor is learned directly during training. 
Moreover, the learned descriptors are optimal as functions for finding functional correspondences between shapes which can be transformed into point correspondences.
Since our method is based on this framework, we give a more detailed introduction in Sec.~\ref{sec:method}. 
The first version of this network, as it was introduced in \cite{litany2017deep}, is trained using ground truth correspondences between shapes.  

Groueix \etal \cite{groueix20183d} proposed to learn correspondences and encoding of 3D shapes together. Their \textit{Shape Deformation Networks} take the input shape and try to align a template with it.
For finding correspondences between pairs of shapes, they deform the template twice, and extract the correspondence through the template. 
This requires a suitable template and ground-truth correspondences for training.
However, obtaining ground truth dense correspondences between shapes is expensive, since it requires either manual labeling via human annotators or special capturing setups, like \cite{bogo2014faust}. 
To overcome the need for labeled data, Halimi \etal \cite{halimi2019unsupervised} proposed to use geodesic distance matrices as supervisor signals during training. 
They used FM-NET \cite{litany2017deep} to obtain a functional map between shapes, and tried to minimize the geodesic distance distortion between the resulting correspondences instead of relying on ground-truth error. 
Our method follows a similar methodology, but we advocate for heat kernels instead of geodesic distances as the supervisor signal.

In parallel to \cite{halimi2019unsupervised}, Roufosse \etal \cite{roufosse2019unsupervised} proposed another unsupervised learning method for learning dense correspondences using FM-NET. 
They used the functional maps itself as the supervisor signal. 
The idea is based on putting constraints on functional maps, such as bijectivity, orthogonality and Laplacian commutativity which are assumed to be properties of the optimal map. 
They predict the functional map between shapes in both directions (source-to-target and target-to-source) while preserving the aforementioned properties in the best possible way. 

\subsection{Heat Kernels}
Heat kernels are a popular tool in shape analysis and have been utilized for many applications in 3D correspondence \cite{coifman2005geometric,ovsjanikov10onepoint,rodola19fmpm}. They have also been used as an approximation of adjacency matrices in \cite{hu2013spectral}. Moreover, they are the basis for the famous Heat Kernel Signature \cite{sun2009concise} and can provide a good replacement for Gaussian kernels \cite{vestner2017efficient}. 
It was also shown in \cite{vestner2017efficient} that using heat kernels gives the bistochistic relaxation of the Quadratic Assignment Matching beneficial properties for optimization. In this paper, we also use heat kernels as a pair-wise descriptor. We explain how we calculate heat kernels on shape and how we use them for training correspondence networks in Sec.~\ref{sub:heatkernels} and Sec.~\ref{sec:method}.

\begin{figure*}
\begin{center}
\includegraphics[width=0.99\linewidth]{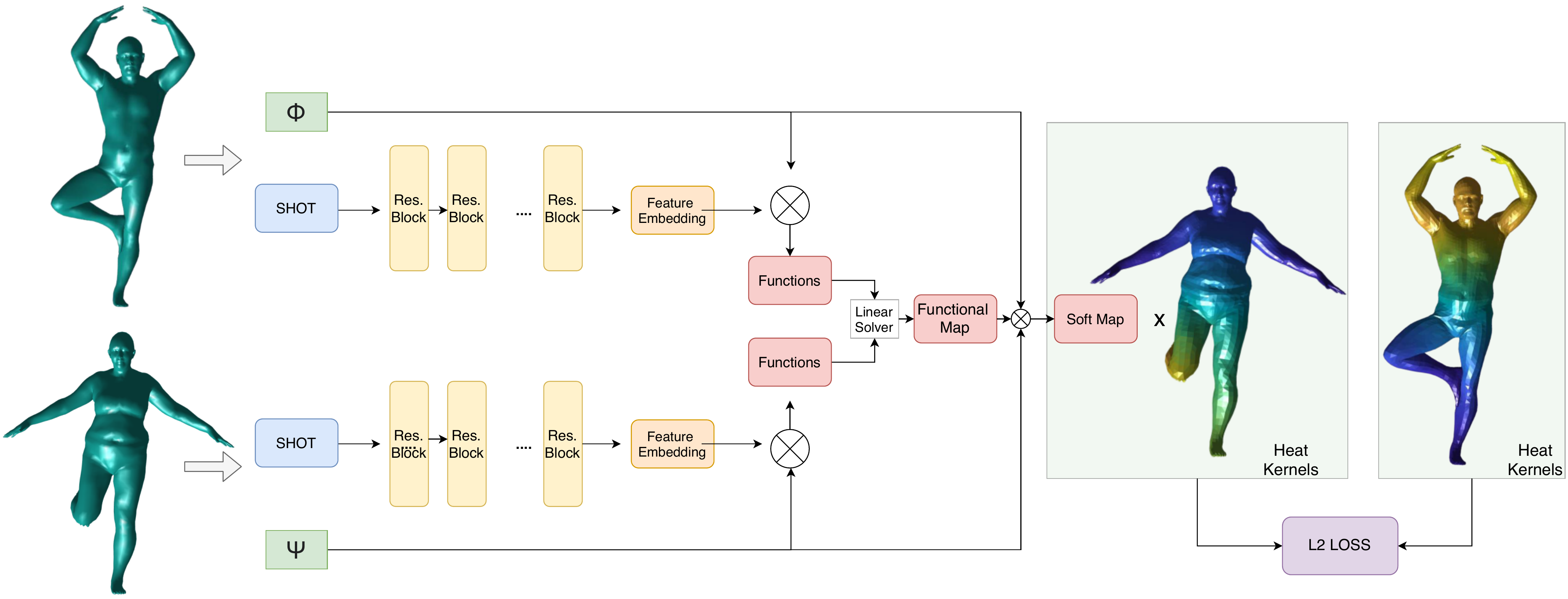}
\end{center}
   \caption{Overview of our proposed network. The network extracts feature embeddings from their SHOT descriptors. Afterwards, the features are projected onto the eigenfunctions and, using a linear solver, the functional map and the soft correspondence map are calculated. The network is trained using the distortion between heat kernels as the supervisor signal.}
\label{fig:main}
\end{figure*}

\section{Background} \label{sec:background}
This section will introduce the necessary background to understand the rest of the paper.
We model shapes as Riemannian 2-manifolds $\mathcal{X}$  with a distance function $d_\mathcal{X}:\mathcal{X}\times\mathcal{X}\to\mathbb{R}$ and define $\pi:\mathcal{X}\to\mathcal{Y}$ as an {\em isometry} map which satisfies
\begin{equation}\label{eq:isometry}
    d_\mathcal{X}(x_1,x_2) = d_\mathcal{Y}({\pi}(x_1),{\pi}(x_2))\,.
\end{equation}
for any pair $x_1,x_2\in\mathcal{X}$.
The correspondence problem for isometric pairs tries to find a map ${\pi}$ which fulfills the distance criterion defined in Eq.~\eqref{eq:isometry}. The optimization aims to minimize the distortion error:
\begin{equation}\label{eq:distortion}
    L(\pi) =  \sum_{  \substack{x_1,x_2 \in \mathcal{X}}}
    \hspace{-0.4cm}
    \left( d_\mathcal{X}(x_1,x_2) - d_\mathcal{Y}({\pi}(x_1),{\pi}(x_2)) \right)^2.
\end{equation}

Generally, the geodesic distance describes the length of the shortest path between two vertices on the surface.

\subsection{Functional Maps}\label{subsec:fm}
The idea of functional maps is to replace finding an point-to-point correspondence $\pi$ between $\mathcal{X}\to\mathcal{Y}$ with seeking a functional map $\mathcal{T}$, which represents a correspondence between functions on the surfaces, namely $\mathcal{F}(\mathcal{X})\to \mathcal{F}(\mathcal{Y})$ \cite{ovsjanikov2012functional}. 
This can be shown to be a linear mapping and given the right basis for the function spaces $\mathcal{F}(\mathcal{X}), \mathcal{F}(\mathcal{Y})$ formulated as a very low dimensional problem in comparison the point-to-point correspondence. 
The basis proposed by \cite{ovsjanikov2012functional}, and chosen by basically every subsequent work, are the eigenfunctions of the Laplace-Beltrami operator (LBO), because they are invariant under isometries and frequency ordered.

\cite{litany2017deep} combines deep learning methods and functional maps to solve the dense shape correspondence problem. 
Instead of depending on hand-crafted descriptors as a functions to guide the functional map optimization, SHOT descriptors \cite{tombari2010unique} are processed with a multi-layer perceptron network, and a new feature representation is learned during training. 
In forward time SHOT descriptors are processed with fully connected layers wherein their dimension is preserved. After, they are projected onto the eigenbasis, and point correspondences are extracted using functional maps. Finally, the loss is calculated with the aim of optimizing distortion error (see Eq.~\ref{eq:distortion}), and the error is propagated while training the network.

\subsection{Heat Kernels}
\label{sub:heatkernels}
\begin{figure}
\begin{center}
\includegraphics[width=0.85\linewidth]{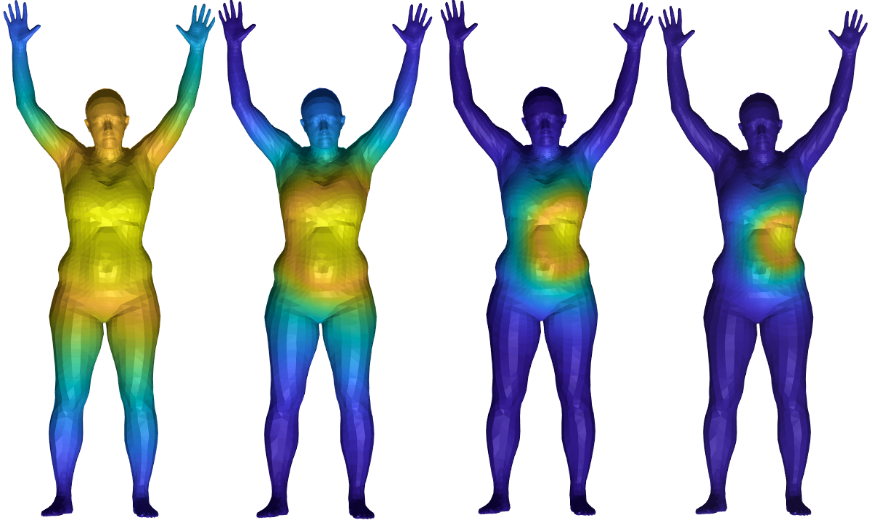}
\end{center}
   \caption{Visualisation of solutions of heat diffusion equation with different diffusion times ($0.1$, $0.03$, $0.01$, $0.003$ from left to right). Yellow are high values, blue is zero.}
\label{fig:short}
\end{figure}

On a manifold $\mathcal{X}$ the {\em heat diffusion equation}  is 
\begin{align}
 \frac{\partial u(t,x)}{\partial t} &= \Delta_{\mathcal{X}}  u(t,x)\,,
\end{align}
with the initial condition $u(0,x) = u_0(x)$ and additional boundary conditions if applicable. 
The equation describes how a given initial heat distribution $u_0$ on $\mathcal{X}$ diffuses over time $t$. 
$u: [0,\infty) \times \mathcal{X} \rightarrow \mathbb{R}$ represents the amount of heat at point $x$ at time $t$. 
In more informal terms, we can imagine putting a heat source onto a point and letting the heat diffuse over the surface of the shape without anything escaping into the surrounding space. 
If a point is far away from the heat source in terms of geodesic distance, the amount of heat arriving there in a certain time is normally less than to a close point.
See Fig.~\ref{fig:short} for a visualization.

The solution for the heat diffusion equation can be calculated by using the equation below which is linear in the initial distribution:
\begin{align}
 u(t,x) &= \int_\Xx k(t,x,x') u_0(x') dx'\,,
\end{align}
where $k:\Rr^+\times \mathcal{X}\times \mathcal{X} \rightarrow \mathbb{R}$ is the \emph{heat kernel}. The values of $k$ represent the the amount of heat transported from $x'$ to $x$ in time $t$. 
With the help of Laplace-Beltrami eigendecomposition heat kernels can be calculated in closed form using the eigenfunctions and -values:
\begin{align}
    k(t,x,x') &= \sum_{k=0}^\infty e^{-\lambda_k \cdot t} \phi_k(x) \phi_k(x')
    \label{eq:heatkernel}
\end{align}
$\lambda_k, \phi_k$ denote the $k$-th LBO eigenvalue and -function. 
If the sum is restricted to a fixed $k$, this equation can be written as a matrix multiplication.

One caveat of heat kernels is that you need to choose the time parameter $t$ carefully, if it is too big the heat diffuses equally to every point or if its too small numerical issues arise. 
One possible option is choosing $t$ by visual inspection. 
Another option is looking at the variance of heat values of some vertices or all vertices to adjust that parameter automatically without visual inspection.

\section{Method} \label{sec:method}

In this section, we will explain how we construct our loss formulation in an unsupervised way and used heat kernels for curriculum learning. 

\subsection{Network loss}

We use FM-NET which is based on deep functional framework introduced in \cite{litany2017deep}. 
We use the same architecture with fully-connected residual layers and exponential linear units (ELU) \cite{clevert2015fast}. 
See Fig.~\ref{fig:main} for an overview.
The input layer takes SHOT descriptors $S \in \mathcal{R}^{N_{s} \times D}$, where $N_{s}$ is the number of vertices and $D$ is the dimension of SHOT descriptor($352$ in our all experiments). 
All fully-connected layers preserve the dimension which means the output of the network is again $N_{s}\times D$.
Next, the learned point descriptors are projected onto the eigenvectors $\Phi \in \mathcal{R}^{E\times N_{s}}$ ($E$ is number of used eigenvalues) to obtain the final function coefficient representation of the shape $F \in \mathcal{R}^{E\times D}$.
The same process is applied to target shape $T$. $T$ can have a different number of vertices but its function representation $G\in \mathcal{R}^{E\times D}$ has the same dimensions as $F$ due to being projected onto the same number of eigenfunctions. 
Note that both shapes are processed with shared layers as in siamese networks. 
Afterwards, the functional map $C$ is obtained by solving the linear equation
\begin{align}
\bb{G} = \bb{CF}.
\end{align}
By using this functional map formulation instead of finding point-to-point correspondence we find a lower dimensional correspondence between functions of shapes. 
Instead of solving a linear equation with dimensions $\mathcal{R}^{N\times D}$, the result is in $\mathcal{R}^{E\times D}$, where $N >> E$, which is more efficient since linear solvers generally depend on the Cholesky decomposition with time complexity $ O(n^{3})$.

However, quantifying the error induced by the functional map without knowing the ground-truth map is hard. 
We convert $C$ into a soft-correspondence map $P \in \mathcal{R}^{N_{s} \times N_{t}}$ by transferring indicator functions from $S$ to $T$ using $C$,
\begin{align}
    \bb{P = |\Psi C \Phi^T A|_{\| \cdot \|}}
\end{align}
where $A$ is the mass matrix of the source shape. 
Afterwards, we convert the mapped indicator functions into probability distributions by taking $Q = P \circ P$, where $\circ$ is the Hadamard product. 
The $i^{th}$ row of $j_{th}$ column of $Q$ represents the probability of the $i^{th}$ vertex of the source corresponding to the $j_{th}$ vertex of the target.
We used same unsupervised loss function as Halimi \etal \cite{halimi2019unsupervised} proposed. It is defined as:

\begin{equation}
    \label{unsupervised_loss}
    \ell_{\mathrm{uns}} (\mathcal{X},\mathcal{Y}) =\frac{1}{|\mathcal{X}|^2}  \norm{\bb{D}_{\mathcal{X}} - \bb{Q}^T \bb{D}_{\mathcal{Y}} \bb{Q}   }^2_{\mathrm{F}}
\end{equation}

where $\bb{D}_{\mathcal{X}}$ and $\bb{D}_{\mathcal{Y}}$ represent the geodesic distance matrices on source and target shape.

\paragraph*{Heat Kernel. } Instead of using geodesics distance matrices, we used heat kernels as the pairwise supervisor signal. 
Preserving the geodesic distances, as it is done in \cite{halimi2019unsupervised}, implies that an isometry is found. 
We saw in Eq.~\eqref{eq:heatkernel} that heat kernels can be calculated using the eigenfunctions and values of the Laplace-Beltrami operator only which implies that heat kernel are also invariant under isometries.
However, geodesic distances assume large values in points far away from each other whereas heat kernel achieve their maximal values on the point itself and its immediate neighborhood. 
In the case of approximate isometries, due to noise or different classes, a change on one vertex will affect not only its neighborhood but geodesic distances on the entire shape. 
Due to the Frobenius norm in Eq.~\eqref{unsupervised_loss} a single large error will have a lot more influence on the loss than a collection of small errors. 
This will lead to many local distortions when the isometry assumption is violated, see Fig.~\ref{fig:partial_vis}.
The influence of heat kernels can be controlled via the diffusion time and for small times the values for further points are nearly zero which means large outliers do not dominate the loss.
Additionally, heat kernels are computationally more efficient compared to geodesic distances since they can be calculated in a closed form, as explained in Sec.~\ref{sub:heatkernels}.
The drawback is that the global information of geodesic distances guides the optimization to the correct local optimum whereas more local information is prone to get stuck in unfortunate local minima. 
To counteract this behavior we introduce a curriculum learning approach in the next section, and show in Sec.~\ref{sec:exp} that our results are equal or better to \cite{halimi2019unsupervised}.

\subsection{Curriculum Learning}\label{sec:curriculum}

Curriculum learning is a teaching strategy where in every step the task to solve gets more challenging with the aim of solving a complex task in the end. This learning strategy is used for training deep neural networks and showed effective performance on various problems \cite{hacohen2019power,bengio2009curriculum,corman2014supervised}.
As discussed in Sec.~\ref{sub:heatkernels} we can control the locality of the heat diffusion via the diffusion time parameter which determines how far the heat will spread. 
For instance, as can be seen in Fig.\ref{fig:short}, higher diffusion values lead to a more spread out and global solution.
When the time parameter is small, the information is very localized and only solutions in which neighborhoods are well preserved give a good loss.
Unfortunately, this makes the loss function harder to optimize due to stronger local optima.
We use this property and start training with higher diffusion times to get good gradient information. Then, we gradually decrease the time when we are already in the vicinity of the optimal solution to get a more precise placement.

\begin{figure}
\centering

\definecolor{mycolor1}{rgb}{0.00000,0.44700,0.74100}%
\definecolor{mycolor2}{rgb}{0.85000,0.32500,0.09800}%
\begin{tikzpicture}

\begin{axis}[%
width=.8\linewidth,
height=.45\linewidth,
at={(0.797in,0.617in)},
scale only axis,
xmin=0,
xmax=30,
ymin=-5,
ymax=15,
xmajorgrids,
ymajorgrids,
every x tick label/.append style={font=\color{black}, font=\footnotesize},
every y tick label/.append style={font=\color{black}, font=\footnotesize},
axis background/.style={fill=white},
axis x line*=bottom,
axis y line*=left,
x label style={at={(axis description cs:0.5,0.02)},anchor=north},
y label style={at={(axis description cs:0.1,.5)},rotate=0,anchor=south},
xlabel={\footnotesize number of vertices(x1000)},
ylabel={\footnotesize time (second) in log},
x tick label style={/pgf/number format/fixed},
legend style={at={(0.575,0.02)},anchor=south west,legend cell align=left,align=left,draw=white!15!black,font=\tiny},
]

\addplot [color=mycolor1, line width=1.25pt]
  table[row sep=crcr]{%
0.412	2.91993056011912\\
0.453	3.11151330654132\\
0.499	3.20700565227912\\
0.549	3.40784192436627\\
0.604	3.52562496003478\\
0.665	3.59936494691148\\
0.732	3.78236979719279\\
0.805	3.87743156065004\\
0.885	4.1953204795718\\
0.974	4.47346569499228\\
1.071	4.56840246179749\\
1.18	4.77068462446021\\
1.298	5.08913835559148\\
1.427	5.14306608127584\\
1.569	6.01112010476376\\
1.726	6.14422855166123\\
1.898	6.3611124767961\\
2.087	6.66265365406547\\
2.297	6.88369309262967\\
2.526	7.11078591486698\\
2.78	7.47607592659415\\
3.058	7.64717994576835\\
3.364	7.68238105890574\\
3.699	7.9011070469934\\
4.069	8.13082899387045\\
4.475	8.34969127922025\\
4.921	8.55005720503558\\
5.413	8.72758700920367\\
5.954	9.03894807267025\\
6.55	9.16907557109781\\
7.205	9.43641562165993\\
7.926	9.66603841248713\\
8.717	9.89501047341473\\
9.586	10.1293230772367\\
10.543	10.3618297002454\\
11.595	10.6196271206348\\
12.753	10.8933569450747\\
14.026	11.0887558633964\\
15.429	11.3257627084568\\
16.969	11.5915275200173\\
18.664	11.9685182550703\\
20.528	12.1389177981272\\
22.577	12.4530232662376\\
24.831	12.6519564028258\\
27.31	12.9561707670202\\
27.894	13.0196337313056\\
};
\addlegendentry{\footnotesize Geodesic Dist.}

\addplot [color=mycolor2, line width=1.25pt]
  table[row sep=crcr]{%
0.412	-1.82\\ 
0.453	-1.60\\ 
0.499	-1.60\\ 
0.549	-1.52\\ 
0.604	-1.52\\ 
0.665	-1.46\\ 
0.732	-1.35\\ 
0.805	-1.26\\ 
0.885	-1.15\\ 
0.974	-1.07\\ 
1.071	-0.98\\ 
1.18	-0.94\\ 
1.298	-0.78\\ 
1.427	-0.78\\ 
1.569	-0.53\\ 
1.726	-0.46\\ 
1.898	-0.40\\ 
2.087	-0.39\\ 
2.297	-0.33\\ 
2.526	-0.26\\ 
2.78	-0.17\\ 
3.058	-0.16\\ 
3.364	-0.16\\ 
3.699	-0.15\\ 
4.069	-0.11\\ 
4.475	-0.07\\ 
4.921	-0.01\\ 
5.413	0.03\\ 
5.954	0.05\\ 
6.55	0.09\\ 
7.205	0.18\\ 
7.926	0.26\\ 
8.717	0.34\\ 
9.586	0.43\\ 
10.543	0.52\\ 
11.595	0.60\\ 
12.753	0.70\\ 
14.026	0.76\\ 
15.429	0.86\\ 
16.969	0.96\\ 
18.664	1.05\\ 
20.528	1.14\\ 
22.577	1.24\\ 
24.831	1.30\\ 
27.31	1.43\\ 
27.894	1.45\\
};
\addlegendentry{\footnotesize Heat}

\end{axis}

\end{tikzpicture}%
   \caption{Comparison of times for calculating heat kernels and geodesic distance matrices with varying number of vertices. Please note that the y-axis is in log scale. }
\label{fig:time_curve}
\end{figure}
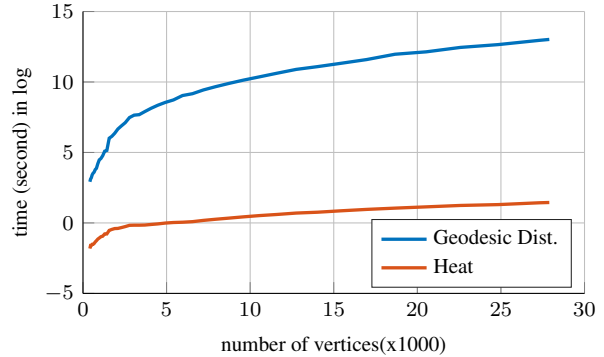

The main idea in curriculum learning settings is that easy examples are optimized first then harder examples are added. 
Hard examples are normally obtained via hard mining which is a time consuming process. However, our heat kernel formulation becomes naturally more challenging, and this allows us to control difficulty without any mining process.
To create such curriculum learning procedure, we start training with heat kernels with a high time parameter and decrease the time after some iterations.    
In our experiments we called this way of training heat decay. 
Moreover, since it is fast to calculate heat kernels, this extra computation step does not lead to longer training times. 
In contrast, geodesic distance matrices need to be pre-calculated before training because of their complexity. We measured calculation times of heat kernel and geodesic distance matrices for shapes that contain different number of vertices. As can be seen in Fig.\ref{fig:time_curve}, calculating heat kernels is extremely fast compared to geodesic distances. 

Our curriculum learning follows the idea of coarse-to-fine optimization in spectral space that has also been explored in \cite{vestner2017efficient} and \cite{melzi2019zoomout}. 
However, both are non-learning methods which means they need to run multiple iterations during inference time. 
Our method only applies this strategy during training, and the run time of inference does not suffer from additional calculations.

\subsection{Implementation Details}
We implemented our framework in Tensorflow \cite{abadi2016tensorflow} based on the original Deep Functional Maps implementation \footnote{https://github.com/orlitany/DeepFunctionalMaps}. 
For all shapes, we extracted 150 eigenvectors and $352$-dimensional SHOT descriptors. 
We calculate SHOT descriptors using 10 bins and a radius equal to $5\% $ of total area of the shape.
We pre-calculated all heat kernels using eigenvectors and eigenvalues before training. 
If the shapes are too big for training (\textasciitilde$100k$) we down-sampled them via edge contraction \cite{garland1997surface} to \textasciitilde$7k$ vertices. 
However, during testing we used the original resolution. We did not use any post processing steps for refining the correspondences.
We use the Adam \cite{kingma2014adam} optimizer with an initial learning rate of $0.001$ for all of our training. In our experiments, using fancy learning rate schedules did not improve our results significantly.

Further, for choosing the decay values and intervals, we used a validation set. For calculating validation loss, we calculated distortion error using heat kernels with initial time diffusion value. However, on most benchmarks we observed no additional improvements after decaying the time value twice.

\section{Experiments}
In this section we show that heat kernels can obtain a similar or better performance than geodesic distances while they provide vastly different energy landscape than geodesic distances. 
First, we show that our method is working even under extreme conditions like having only one training example without ground-truth information. 
Then, we try our proposed method on the FAUST Synthetic data set \cite{bogo2014faust} which is a commonly used data set for evaluating shape correspondence performance. 
We also evaluate on the SHREC 16' \cite{cosmo2016shrec, SHREC16-topology} and SHREC 19' data sets which include challenges such as partiality, topological noise and different connectivity respectively. 

We use geodesic error curves as the evaluation measure \cite{kim2011blended}.
This curve contains threshold values on the x-axis and the percentage of correct correspondences on the y-axis. For every vertex on the source shape the difference between the match and the ground-truth match is considered. 
The curve plots the percentage of differences which are lower than the threshold value on the x-axis. The perfect solution would show a constant curve at $1$.

\begin{figure}
     \centering
     \begin{subfigure}[b]{0.23\textwidth}
         \centering
         \includegraphics[height=0.14\textheight]{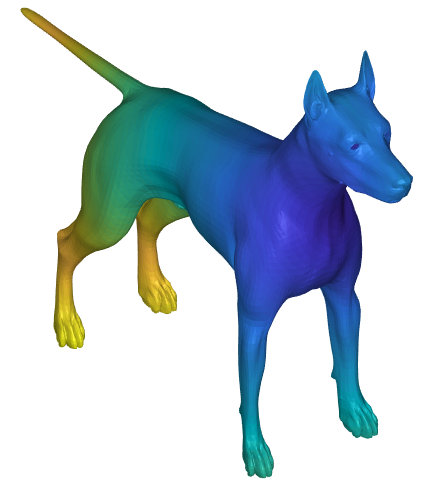}
%         \caption{Geodesic Distance}
         \label{fig:single_ref}
     \end{subfigure}
     \hfill
          \begin{subfigure}[b]{0.23\textwidth}
         \centering
         \includegraphics[height=0.14\textheight]{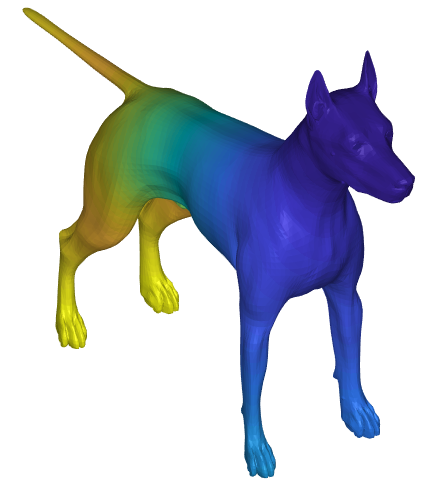}
%         \caption{Heat Kernel}
     \end{subfigure}
     \hfill
          \begin{subfigure}[b]{0.23\textwidth}
         \centering
         \includegraphics[height=0.18\textheight]{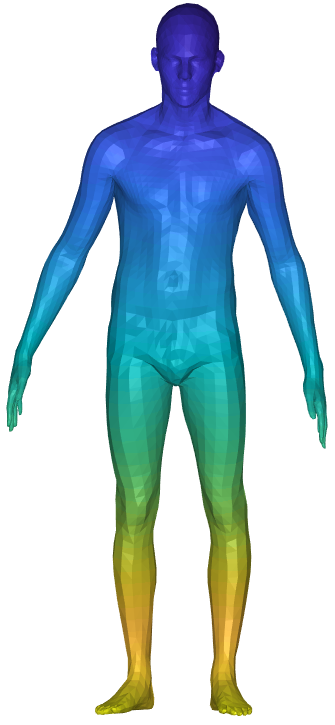}
         \caption{Geodesic Distance}
     \end{subfigure}
     \hfill
          \begin{subfigure}[b]{0.23\textwidth}
         \centering
         \includegraphics[height=0.18\textheight]{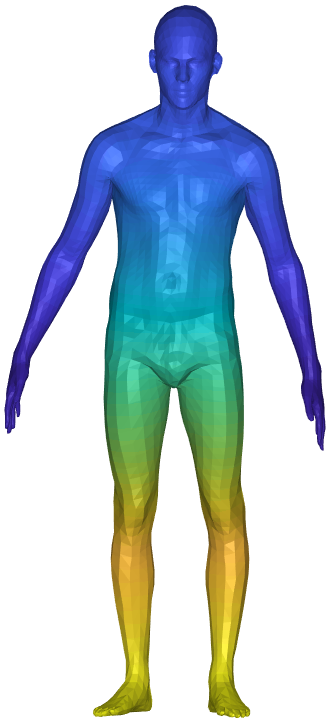}
         \caption{Heat Kernel}
     \end{subfigure}
        \caption{Visualisation of heat kernels and geodesic distances on FAUST \cite{bogo2014faust} and SHREC 16' \cite{cosmo2016shrec} data sets. While blue represent close areas, yellow represent far away areas. As can be seen on both shapes, both heat kernel and geodesic distances provide similar information about closeness.}
        \label{fig:geovsheat}
\end{figure}

\label{sec:exp}
\subsection{Single Shape}

One of the most extreme case in learning for shape correspondences is not having any training examples or other priors. 
We repeat the experiment from Halimi \etal \cite{halimi2019unsupervised} and calculate a correspondence for a single shape pair.
The result and a comparison can be seen in Fig.~\ref{fig:single_exp}.
We used $150$ LBO eigenfuctions, the SHOT descriptor and optimize our network using heat kernels.
Our method achieved comparable, in some parts even better results, than the geodesic distance version and other methods. 
Since no ground truth is available, there is no quantitative evaluation.

\subsection{FAUST - Synthetic data set}

\begin{figure}
     \centering
     \begin{subfigure}{0.15\textwidth}
         \centering
         \includegraphics[height=0.18\textheight]{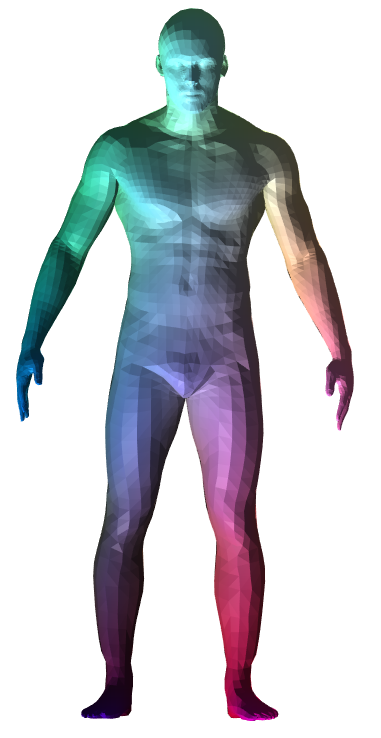}
         \caption{Source Shape}
     \end{subfigure}
     \hfill
          \begin{subfigure}{0.15\textwidth}
         \centering
         \includegraphics[height=0.18\textheight]{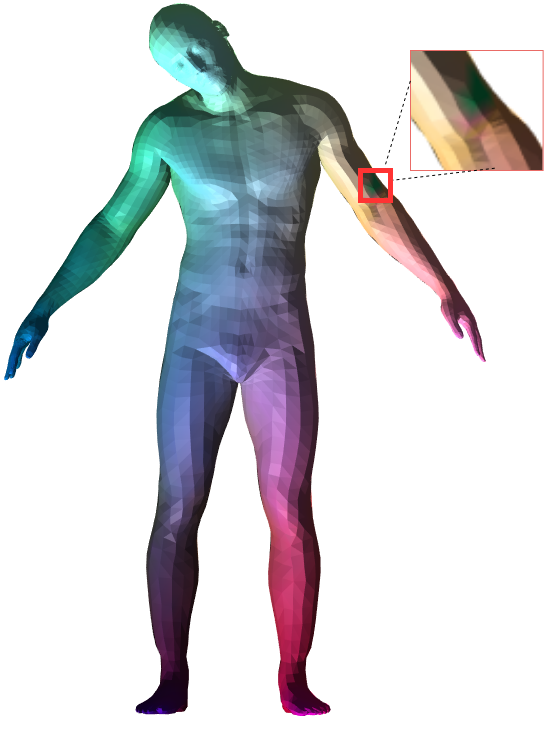}
         \caption{Heat}
     \end{subfigure}
     \hfill
          \begin{subfigure}{0.15\textwidth}
         \centering
         \includegraphics[height=0.18\textheight]{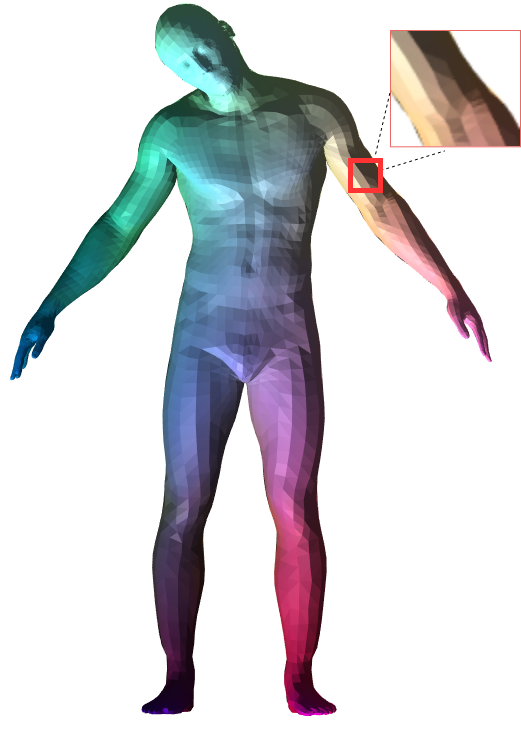}
         \caption{Heat /w decay}
     \end{subfigure}
        \caption{Visualisation of correspondences on a FAUST \cite{bogo2014faust} shape. Same colors represent the correspondences. While the model trained with heat kernel produce very good result, on the left arm there is a small error. However, the model with decayed training did not generate a similar error, since it trained with refined heat kernels also. Please note that, the person in the shape was not included in the training.}
        \label{fig:faust_syn_vis}
\end{figure}

The FAUST data set \cite{bogo2014faust} contains 100 shapes which show 10 different people in 10 different poses. 
The original scans are registered to a template with $6890$ vertices, and we use these registrations. 
For training, we used the first eight persons and all of their poses. We used the remaining the two persons for testing our method. We used $120$ LBO eigenfunctions, and $352$-dimensional SHOT descriptor as the input to our model. 
We evaluate three different unsupervised signals: the geodesic distance matrix as proposed in Halimi \etal \cite{halimi2019unsupervised}, heat kernel with a fixed temperature of $t=0.1$ and decayed heat training with initially $t=0.1$ and later $t=0.01$. When we use the geodesic distance matrix, we train our method with $10k$ iterations similar to \cite{halimi2019unsupervised}. For heat kernels, we also trained our model with $10k$ iterations and decayed the heat kernel values after $5k$ iterations.

\begin{figure}
\centering
\input{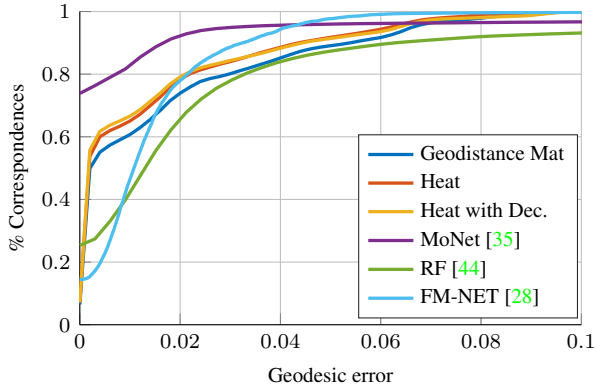}
   \caption{Comparison of heat kernel and geodesic distance matrices on FAUST Synthetic shapes. Training with Heat Kernels and its decayed version obtains better results than with the geodesic distance matrix.}
\label{fig:faust_curve}
\end{figure}

During training, we shuffled vertices of all shapes because the ground-truth is one-to-one, and the network could just learn to return the identity.  
As can be seen in Fig.\ref{fig:faust_curve} heat kernels obtained better results compared to geodesic distances. 
Our interpretation is that decayed heat kernels provide different details on each scale which helps the network to refine the solution. 
Moreover, heat kernels provide similar distance preservation criteria as geodesic distances. 
We visualised heat kernel and geodesic distance on FAUST shape to validate this claim, which can be seen on Fig.\ref{fig:geovsheat}.   

\subsection{SHREC 16' - Partiality}

The SHREC 16' partiality data set \cite{cosmo2016shrec} contains shapes from different classes like dogs and humans. The main challenge of this data set is that the shapes are partial, i.e contain holes or cuts. In our experiment, we choose the dog class with holes similar to \cite{halimi2019unsupervised}.
The data set also has a null shape for each category and correspondences are calculated to that shape for evaluation. In the dog category there are $10$ training and $26$ test shapes.
For each shape we extracted $120$ LBO eigenbases and SHOT descriptors. We trained our models using  i) fixed heat kernel value ($t=1.0$), ii) dynamic heat kernel value ($t=1.0$ then $t=0.1$), and iii) geodesic distance matrix. We used FM-NET with $5$ layers since this data set is small compared to other data sets and trained the model with only $1k$ iterations. 

\begin{figure}[h]
%\begin{center}
%\includegraphics[width=0.99\linewidth]{images/shrec16.pdf}
%\end{center}
\centering
\input{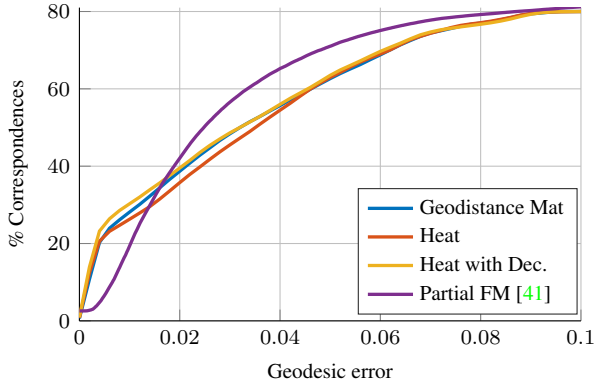}
   \caption{Comparison of heat kernel and geodesic distance matrices on SHREC’16 benchmark for the dog class. Shapes contains holes and this lead to partial matching of deformable shapes. Heat kernel with decayed training obtained comparable results with the geodesic distances.}
\label{fig:partialty_curve}
\end{figure}

\begin{figure*}
\begin{tabular}{p{1.0in}p{1.0in}p{1.3in}p{1.3in}p{1.3in}}
\multicolumn{2}{l}{\multirow[l]{2}{*}{\includegraphics[width=1.8in]{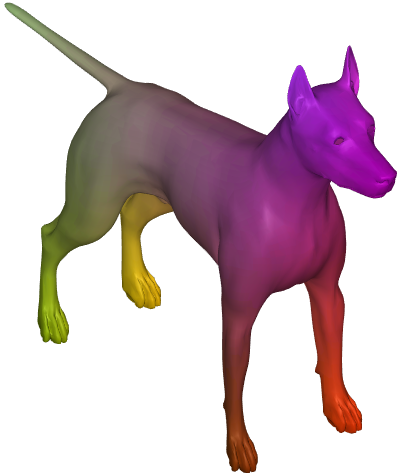}}}  \cr \noalign{\vrule} 
& & \subcaptionbox{\label{1st-fig}Heat}{\includegraphics[width=1.0in, height=1.0in]{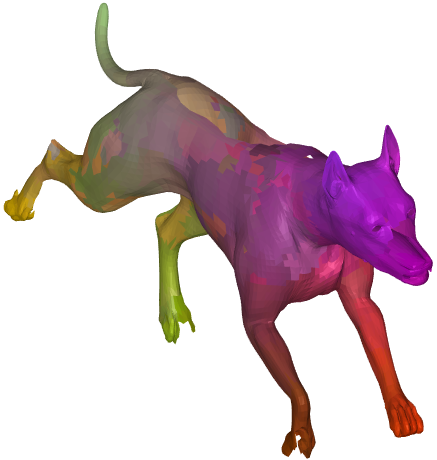}} &
    \subcaptionbox{\label{1st-fig}Heat /w decay}{\includegraphics[width=1.0in, height=1.0in]{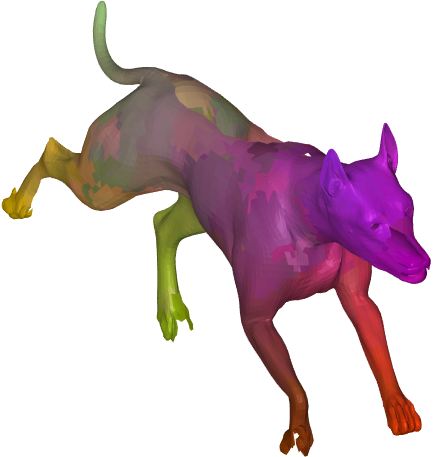}} &
    \subcaptionbox{\label{1st-fig}Geodesic\cite{halimi2019unsupervised}}{\includegraphics[width=1.0in, height=1.0in]{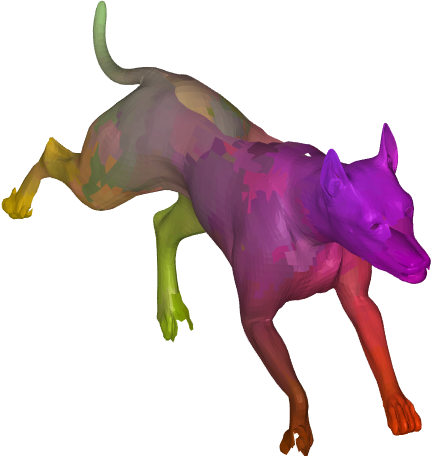}}  \\
& & \subcaptionbox{\label{2nd-fig}Heat}{\includegraphics[width=1.0in, height=1.0in]{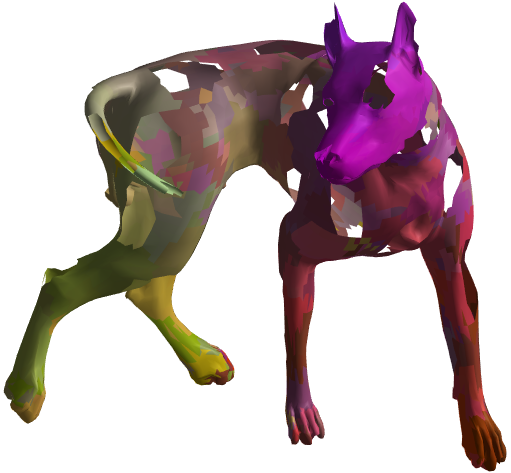}} &
     \subcaptionbox{\label{2nd-fig}Heat /w decay}{\includegraphics[width=1.0in, height=1.0in]{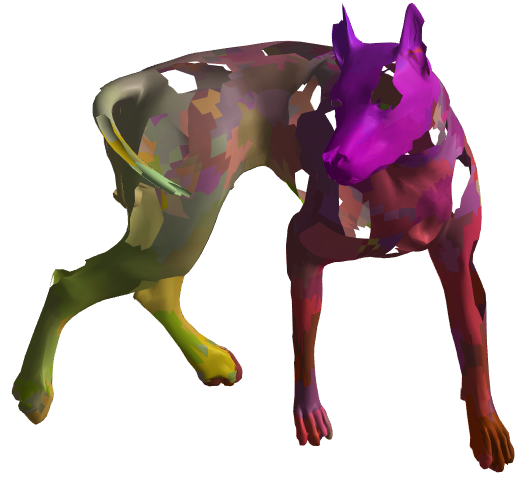}} &
      \subcaptionbox{\label{2nd-fig}Geodesic\cite{halimi2019unsupervised}}{\includegraphics[width=1.0in, height=1.0in]{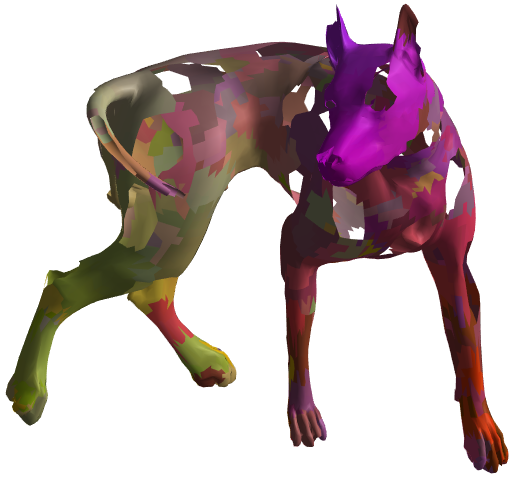}} \\
\end{tabular}
  \caption{Visualisation of some correspondences on SHREC 16 test set. The shape on the left is the source shape. (a) \& (d) is obtained using the model which trained using a fix heat kernel, (b) \& (e) with decayed heat kernel. While test shapes are not complete and in different poses, our method is able to find reasonable correspondences which is similar or better than \cite{halimi2019unsupervised}.}
  \label{fig:partial_vis}
\end{figure*}

As illustrated in Fig.\ref{fig:partialty_curve}, while fixed heat kernels obtain lower scores than geodesic distance matrix, heat kernels with decayed training obtain better scores. In addition to the correspondence error curve we visualised correspondences on two challenging poses in Fig.\ref{fig:partial_vis}.

\subsection{SHREC 16' - Topological Noise}

The SHREC 16' - Topological Noise data set \cite{SHREC16-topology} contains shapes from KIDS data set \cite{rodola2014dense} with topological merging. 
The data set contains $16$ training and $9$ test shapes. For training we used $150$ LBO eigenbases and the original resolution of shapes. During evaluation we looked all pairs of test shapes and calculate similar plot as in previous sections. We compare our method with geodesic distances. 

\begin{figure}
\centering

\definecolor{mycolor1}{rgb}{0.00000,0.44700,0.74100}%
\definecolor{mycolor2}{rgb}{0.85000,0.32500,0.09800}%
\definecolor{mycolor3}{rgb}{0.92900,0.69400,0.12500}%
\definecolor{mycolor4}{rgb}{0.49400,0.18400,0.55600}%
\definecolor{mycolor5}{rgb}{0.46600,0.67400,0.18800}%
\definecolor{mycolor6}{rgb}{0.30100,0.74500,0.93300}%

\begin{tikzpicture}

\begin{axis}[%
width=.8\linewidth,
height=.5\linewidth,
at={(0.797in,0.617in)},
scale only axis,
xmin=0,
xmax=0.25,
ymin=0,
ymax=80,
xmajorgrids,
ymajorgrids,
every x tick label/.append style={font=\color{black}, font=\footnotesize},
every y tick label/.append style={font=\color{black}, font=\footnotesize},
axis background/.style={fill=white},
axis x line*=bottom,
axis y line*=left,
x label style={at={(axis description cs:0.5,0.02)},anchor=north},
y label style={at={(axis description cs:0.1,.5)},rotate=0,anchor=south},
xlabel={\footnotesize Geodesic error},
ylabel={\footnotesize \% Correspondences},
x tick label style={/pgf/number format/fixed},
legend style={at={(0.555,0.02)},anchor=south west,legend cell align=left,align=left,draw=white!15!black,font=\tiny},
]
\addplot [color=mycolor1, line width=1.25pt]
  table[row sep=crcr]{%
0	2.59186656367258\\
0.01	3.38915028136611\\
0.02	4.76359272323923\\
0.03	6.52470939909646\\
0.04	8.66421081963232\\
0.05	11.8310706659159\\
0.06	16.7404270618203\\
0.07	21.1118311346637\\
0.08	24.6987633481227\\
0.09	28.2922523997457\\
0.1	31.8255472986559\\
0.11	35.0297464125853\\
0.12	38.062761187677\\
0.13	41.2638983515032\\
0.14	44.2139106471693\\
0.15	46.8414323947308\\
0.16	49.0553276500261\\
0.17	51.3558860281895\\
0.18	53.692981145922\\
0.19	55.65570627831\\
0.2	57.6074265022789\\
0.21	59.3695124462739\\
0.22	61.1769616789454\\
0.23	62.8190978126739\\
0.24	64.3191684829309\\
0.25	65.6390002646242\\
};
\addlegendentry{ \footnotesize EM \cite{SHREC16-topology}}

\addplot [color=mycolor2, line width=1.25pt]
  table[row sep=crcr]{%
0	0.445294555428997\\
0.01	1.31245446341963\\
0.02	4.51000462883511\\
0.03	9.08664672498074\\
0.04	14.3987000966751\\
0.05	19.9703394871348\\
0.06	25.5057257828931\\
0.07	30.9008892302496\\
0.08	35.7647952028367\\
0.09	40.0027260342255\\
0.1	43.7512922427831\\
0.11	47.1431314799796\\
0.12	50.2471758130442\\
0.13	53.06279522252\\
0.14	55.6360768869578\\
0.15	58.0082396103639\\
0.16	60.1506136802062\\
0.17	62.1230959585937\\
0.18	64.0208762391739\\
0.19	65.833213079631\\
0.2	67.5642360581011\\
0.21	69.2517508285218\\
0.22	70.8398579735419\\
0.23	72.3378073837109\\
0.24	73.7134272564856\\
0.25	75.0557555225489\\
};
\addlegendentry{ \footnotesize GE\cite{burghard2017embedding}}

\addplot [color=mycolor3, line width=1.25pt]
  table[row sep=crcr]{%
0	6.69306308296569\\
0.01	9.75182598172478\\
0.02	16.0739754499995\\
0.03	21.9105202493059\\
0.04	27.0985167507045\\
0.05	31.6650974236389\\
0.06	35.7293417273631\\
0.07	39.3590888180227\\
0.08	42.5098297810401\\
0.09	45.2774398987196\\
0.1	47.7124888384332\\
0.11	49.9247027830963\\
0.12	51.8717109345963\\
0.13	53.5816911524929\\
0.14	55.2548294222688\\
0.15	56.7746062699835\\
0.16	58.2054164273659\\
0.17	59.547385848875\\
0.18	60.8408060035973\\
0.19	62.0718751224272\\
0.2	63.1909999291417\\
0.21	64.3272975452623\\
0.22	65.5001824431898\\
0.23	66.6245064393347\\
0.24	67.8089497477715\\
0.25	68.9986900611532\\
};
\addlegendentry{ \footnotesize RF \cite{rodola2014dense}}
\addplot [color=mycolor4, line width=1.25pt]
  table[row sep=crcr]{%
0	1.04248157168581\\
0.01	11.458563665662\\
0.02	22.15783369188\\
0.03	29.7068572515576\\
0.04	34.9844746021365\\
0.05	38.8725263252066\\
0.06	41.8843979558537\\
0.07	44.29576297842\\
0.08	46.4400699662273\\
0.09	48.4134460152458\\
0.1	50.2166033473378\\
0.11	51.9257367029798\\
0.12	53.5830712367722\\
0.13	55.2076980674985\\
0.14	56.7729465773796\\
0.15	58.2286926173562\\
0.16	59.6300967667798\\
0.17	60.8970953234437\\
0.18	62.1391026961152\\
0.19	63.4394526404186\\
0.2	64.6559444280421\\
0.21	65.888297717007\\
0.22	67.136889121875\\
0.23	68.3636062138447\\
0.24	69.5898515103151\\
0.25	70.8382995954699\\
};
\addlegendentry{ \footnotesize Heat}

\addplot [color=mycolor5, line width=1.25pt]
  table[row sep=crcr]{%
0	1.51699612173882\\
0.01	13.373025924107\\
0.02	24.957181283179\\
0.03	32.2264176389231\\
0.04	36.9584107571043\\
0.05	40.4090579067775\\
0.06	43.0214868622063\\
0.07	45.086694515748\\
0.08	46.8672115434239\\
0.09	48.4942045648836\\
0.1	50.0252548325872\\
0.11	51.4394189810505\\
0.12	52.7864337714567\\
0.13	54.2362868738784\\
0.14	55.7087404949857\\
0.15	57.1349334225867\\
0.16	58.4707071263715\\
0.17	59.6948451523403\\
0.18	60.8871633781558\\
0.19	62.1039608001819\\
0.2	63.3062538308432\\
0.21	64.5120152992734\\
0.22	65.7182938185087\\
0.23	66.9461909703893\\
0.24	68.1265558392719\\
0.25	69.291097743548\\
};
\addlegendentry{ \footnotesize Geodistance Mat}

\addplot [color=mycolor6, line width=1.25pt]
  table[row sep=crcr]{%
0	1.55511864037585\\
0.01	13.6185561757844\\
0.02	25.2222275340859\\
0.03	32.6198719079627\\
0.04	37.4886992295223\\
0.05	40.9517161295123\\
0.06	43.5652359890713\\
0.07	45.631613202808\\
0.08	47.5223919147378\\
0.09	49.2576004880127\\
0.1	50.8988289504286\\
0.11	52.3760462794881\\
0.12	53.7195133670132\\
0.13	55.112836291181\\
0.14	56.5375861750848\\
0.15	57.9556331381031\\
0.16	59.2973282417797\\
0.17	60.4960274804917\\
0.18	61.6676551520274\\
0.19	62.8378596238769\\
0.2	63.9927821723167\\
0.21	65.1573200730187\\
0.22	66.335655053864\\
0.23	67.50483396009\\
0.24	68.7310071916096\\
0.25	69.9027305681487\\
};
\addlegendentry{ \footnotesize Heat with Dec.}

\end{axis}

\end{tikzpicture}%
   \caption{Comparison of Heat Kernels and Geodesic distance matrix on SHREC’16 topological noise benchmark. Shapes contains topological artifacts around their body.}
\label{fig:topnoise_plot}
\end{figure}
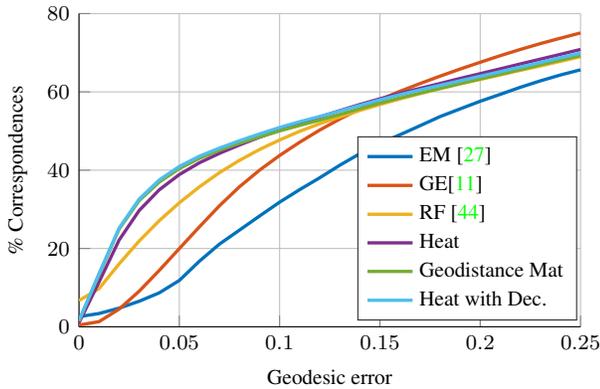

The results of our experiment can be seen in Fig.~\ref{fig:topnoise_plot}. While decayed heat kernels achieved better results than geodesic distance with a big error margin, the decayed version obtained better result on small error margin. 
This show that the model which was trained with decayed heat kernels was able to find correspondences on finer details while fixed heat kernels obtained more rough matches.

\subsection{SHREC 19' - Different Connectiviy}

The Shrec'19 Matching Humans with Different Connectivity challenge \cite{melzi19connectivity}  is designed to evaluate correspondence methods under different density (5k to 50k) and different meshing distribution (uniform and nonuniform). 
The challenge contains different human shapes from variety of data sets like FAUST \cite{bogo2014faust}, TOSCA \cite{bronstein2008numerical}, SCAPE \cite{anguelov2005scape}.

In our experiments, we remeshed all shapes to have about $7k$ vertices due to memory constraints. 
Afterwards, we extracted $150$ LBO eigenfunctions and $352$-dimensional SHOT descriptors from these resized shapes. 
We used FM-NET\cite{litany2017deep} with $7$ residual layers. We trained our models with $10k$ iterations, and decayed after $5k$ iterations.

\begin{figure}
\centering
\input{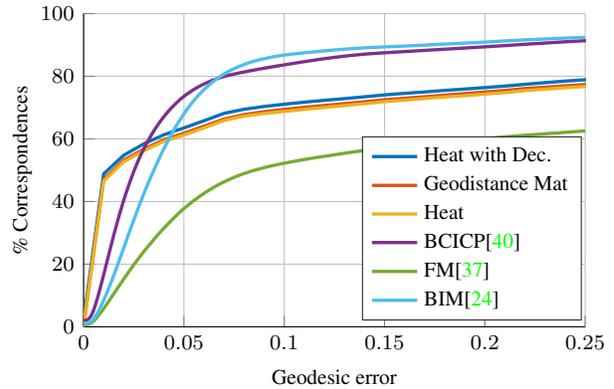}
   \caption{Comparison of Heat Kernels and Geodesic distance matrix on SHREC’19 different connectivity benchmark. Shapes vary in number of vertices and triangles. In this benchmark fixed heat kernel did not outperform geodesic distance matrix, but the decay version did.}
\label{fig:connectiviy_plot}
\end{figure}

During inference we used the original shapes which have high variety on the number of vertices. 
In our experiments we explored three different unsupervised signals:  i) fixed heat kernel value ($t=0.1$), ii) dynamic heat kernel value ($t=0.1$ then $t=0.01$), and iii) geodesic distance matrix. An overview of the results can be seen in Fig.\ref{fig:connectiviy_plot}. As depicted in the figure heat kernel with decay obtain better performance than geodesic distances. 

\section{Conclusion}
\label{sec:conclusion}
We proposed to use heat kernels instead of geodesic distances for training correspondence networks without any ground-truth matches in the FM-Net architecture. 
We show that heat kernel are an equal but more efficient replacement for geodesic distances, even though they provide a less optimal energy landscape for optimization. Networks trained with heat kernels obtain equal or better results on various benchmarks while improving run time significantly. 
Moreover, heat kernel can be recalculated during training, and we proposed a curriculum learning approach with different diffusion times that improved correspondence accuracy on every benchmark. 

\section*{Acknowledgement}
We would like to thank Oshri Halimi for help with using her code. We gracefully acknowledge the support of the Collaborative Research Center SFB-TRR 109 'Discretization in Geometry and Dynamics'. 

{\small
\bibliographystyle{ieee}
\bibliography{egpaper_for_review.bbl}
}

\end{document}